%% file: main.tex
\RequirePackage[svgnames]{xcolor}

\documentclass[11pt,letterpaper]{mystyle}

\usepackage[all]{hypcap}
\usepackage[svgnames]{xcolor}
\usepackage[comma,authoryear,compress]{natbib}
\usepackage{hyperref}[citecolor=lightblue]

\hypersetup{
    colorlinks = true,
    citecolor = {YaleBlue},
}

\usepackage{algorithm}
\usepackage{algorithmicx}
\usepackage{algpseudocode}
\usepackage{microtype}
\usepackage{graphicx}
\expandafter\def\csname ver@subfig.sty\endcsname{}
\usepackage{booktabs} %
\usepackage{float}
\usepackage{bigstrut}

\usepackage{amsmath}
\usepackage{amssymb}
\usepackage{mathtools}
\usepackage{amsthm}
\usepackage{mathrsfs}
\usepackage{nicefrac}
\usepackage{dsfont}
\usepackage{enumitem}
\usepackage{subcaption}
\usepackage{graphicx,subfig}
\usepackage{cleveref}
\usepackage{bxcoloremoji}

\usepackage{float}

\usepackage[utf8]{inputenc} %
\usepackage[T1]{fontenc}    %
\usepackage{hyperref}       %
\usepackage{url}            %
\usepackage{booktabs}       %
\usepackage{amsfonts}       %
\usepackage{nicefrac}       %
\usepackage{microtype}      %
\usepackage{graphicx}
\usepackage{subcaption} 
\usepackage{amssymb}
\usepackage{fdsymbol}
\usepackage{wrapfig}
\usepackage{lipsum}
\usepackage{enumitem}
\usepackage{stackengine}
\usepackage[font=small,labelfont=bf]{caption}
\usepackage{color}
\usepackage{adjustbox}

\usepackage{rotating}
\usepackage{makecell}

\input{macro}

\newtcolorbox{AIbox}[2][]{aibox,title=#2,#1}
\definecolor{lightblue}{rgb}{0.22,0.45,0.70}%
\definecolor{Gray}{gray}{0.95}
\definecolor{Cornsilk}{rgb}{1.0, 0.97, 0.86}

\usepackage{amsmath}
\usepackage{amsthm}  
\usepackage{graphicx}  
\usepackage{tcolorbox} 
\usepackage{multirow}
\usepackage{colortbl}
\usepackage{bbding}
\usepackage{wrapfig}
\usepackage{array}  

\usepackage{amssymb,booktabs}
\usepackage{tabularx}
\usepackage{bbding}
\usepackage{pifont}
\usepackage{makecell}
\usepackage[table]{xcolor}
\usepackage{booktabs}

\usepackage[all]{hypcap}

\usepackage{tcolorbox}
\usepackage{array}
\usepackage{xcolor}
\usepackage{booktabs}  
\usepackage{geometry}  
\usepackage{enumitem}
\title{RemoteZero: Geospatial Reasoning with Zero Labels}

\runningtitle{RemoteZero: Geospatial Reasoning with Zero Labels}

\author{
  Liang Yao$^{1}$,
  Fan Liu$^{1, \dagger}$, 
  Shengxiang Xu$^2$, 
  Chuanyi Zhang$^1$, \\
  Rui Min$^1$,
  Shimin Di$^2$, and
  Yuhui Zheng$^1$
}

\affil[1]{Hohai University}
\affil[2]{Southeast University}

\correspondingauthor{Fan Liu. Email \href{mailto:fanliu@hhu.edu.cn}{fanliu@hhu.edu.cn}}

\begin{document}

\input{sections/abstract}

\maketitle
\vspace{3mm}
\input{sections/introduction}

\input{sections/motivation}

\input{sections/method}
\input{sections/experiments}
\input{sections/relatedwork}
\input{sections/conclusion}
\clearpage
\bibliography{main}
\clearpage
\end{document}

%% file: macro.tex
\newcommand{\x}{\mathbf{x}}

\definecolor{blanchedalmond}{rgb}{1.0, 0.92, 0.8}
\definecolor{carmine}{rgb}{0.59, 0.0, 0.09}
\definecolor{lightblue}{rgb}{0.22,0.45,0.70}%

\renewcommand{\mathbf}{\boldsymbol}

\makeatletter
\def\Ddots{\mathinner{\mkern1mu\raise\p@
\vbox{\kern7\p@\hbox{.}}\mkern2mu
\raise4\p@\hbox{.}\mkern2mu\raise7\p@\hbox{.}\mkern1mu}}
\makeatother

\definecolor{amaranth}{rgb}{0.9, 0.17, 0.31}
\definecolor{antiquebrass}{rgb}{0.8, 0.58, 0.46}
\definecolor{antiquefuchsia}{rgb}{0.57, 0.36, 0.51}
\definecolor{chromeyellow}{rgb}{0.31, 0.47, 0.26}

\newcommand{\github}{\raisebox{-1.5pt}{\includegraphics[height=1.05em]{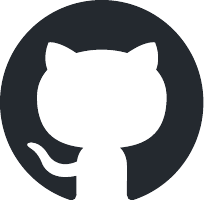}}}

%% file: sections/abstract.tex
\begin{abstract}
Geospatial reasoning requires models to identify image regions that satisfy complex and often implicit user intents. 
Recent reinforcement learning approaches improve reasoning without manually annotated reasoning traces, but still require human-provided target labels to construct rewards, limiting their use on large-scale unlabeled Earth observation data. 
We introduce RemoteZero, a label-free framework for reinforcement-based geospatial reasoning. 
Our key observation is twofold: MLLMs are often more reliable at evaluating candidates than generating solutions, while aerial imagery reduces interference in region-level verification.
Therefore, RemoteZero converts each predicted region into a visual crop and uses its semantic consistency with the query as an intrinsic reward for GRPO optimization. 
This formulation removes the need for human-provided solution labels and further supports iterative self-evolution by reusing previous-round models as verifiers. 
Experiments show that RemoteZero outperforms strong supervised baselines, applies effectively to other Earth observation tasks, and continues to improve through self-evolution as the training data expand. We hope this direction can broadly benefit the Earth observation community.
\vspace{2mm}


\vspace{5mm}

\github{} \textbf{Code Repository}: \href{https://github.com/1e12Leon/RemoteZero}{https://github.com/1e12Leon/RemoteZero}

\coloremojicode{1F4E7} \textbf{Contact}: \href{mailto:yaoliang@hhu.edu.cn}{yaoliang@hhu.edu.cn}

\end{abstract}

%% file: sections/introduction.tex
\vspace{-4mm}
\section{Introduction}

\begin{figure*}[t]
  \centering
  \includegraphics[width=1.0\textwidth]{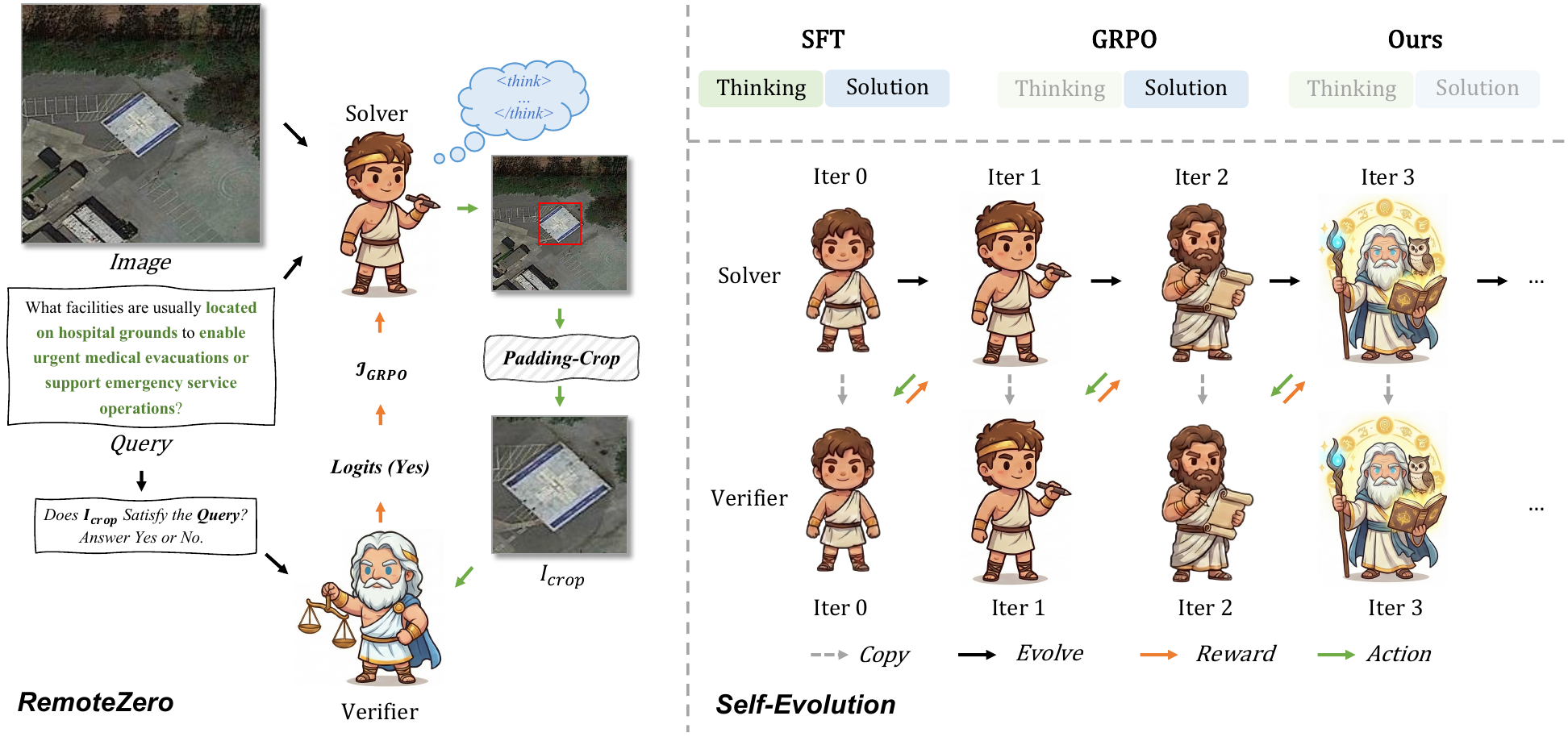}
    \caption{Conceptual overview of RemoteZero. Left: predicted regions are cropped and semantically verified to produce label-free rewards. Top-right: comparison with SFT and conventional GRPO. Bottom-right: previous-round models are reused as verifiers for iterative self-evolution.}
  \label{fig:teaser}
\end{figure*}

The long-term objective of remote sensing foundation models~\cite{xu2025towards,zhang2024vision,zhou2024towards} is to advance Earth science, in which many real-world tasks carry substantial social and economic value.
Geospatial reasoning~\cite{li2025segearth,yao2026remotereasoner} is a key task toward this objective, as it requires models to identify regions that satisfy implicit, non-technical queries rather than predefined object categories. 
For example, after an earthquake, a decision-maker is unlikely to ask merely to ``\textit{Detect a playground}.'' 
A realistic request may instead be to ``\textit{Identify an optimal zone for resettling victims that maximizes capacity while facilitating rapid facility deployment}.'' 
Such a query does not name a specific visual target; it describes a functional need whose solution depends on jointly interpreting the intended use, spatial constraints, and scene evidence. 
The model must infer which visible region best serves the intended function, shifting remote sensing from content recognition toward decision-oriented reasoning.

Recent RFT/GRPO-based studies have advanced tasks adjacent to geospatial reasoning, including referring expression segmentation~\cite{fiaz2025geovlm,zhang2025geor1} and visual grounding~\cite{wang2025geozero,li2026georeason,sun2026geosolver}. 
However, these tasks typically specify the target explicitly, requiring models to identify a given object or referent rather than infer the latent functional intent behind a decision-oriented query.
SegEarth-R1~\cite{li2025segearth} pioneered the Geospatial Pixel Reasoning task, employing Supervised Fine-Tuning (SFT)~\cite{wang2025parameter,han2024parameter} to align models with manually curated triplets of images, reasoning chains, and segmentation masks. However, this paradigm depends on labor-intensive annotations and may overfit to fixed reasoning patterns. 
To reduce the reliance on annotated reasoning traces, RemoteReasoner~\cite{yao2026remotereasoner} introduced Group Relative Policy Optimization (GRPO)~\cite{shao2024deepseekmath} into remote sensing. 
By optimizing the policy through reinforcement learning rather than SFT, it allows MLLMs to construct reasoning chains more autonomously while preserving general capabilities. 
Nevertheless, it still relies on human-annotated bboxes to compute accuracy rewards, such as IoU, limiting its ability to leverage large-scale unlabeled Earth observation data for training.

These limitations motivate us to seek a learning signal that remains observable without human-provided labels. One promising direction is to evaluate the semantic compatibility between a proposed region and the query. This possibility is supported by a general generation--verification asymmetry in MLLMs~\cite{he2026far}: evaluating whether a specified candidate satisfies a requirement is often more reliable than generating a valid structured solution from a large output space. Overhead remote sensing imagery further improves the reliability of such judgments, as limited occlusion and cross-object overlap reduce semantic interference from unrelated targets. 
Consequently, these properties suggest that query--region compatibility can provide a viable label-free learning signal for geospatial reasoning in remote sensing.

Building on this observation, we introduce RemoteZero, a label-free framework illustrated in Fig.~\ref{fig:teaser}. 
RemoteZero employs a Generate-Crop-Verify loop, where each predicted box is converted into a visual crop and evaluated for semantic consistency with the query, providing an intrinsic reward for GRPO without human-labeled coordinates. 
This design enables policy optimization directly on large-scale unlabeled Earth observation data and supports two complementary learning modes: verifier-based distillation from a stronger MLLM and iterative self-evolution using the model's own verification feedback. 
By leveraging the stronger verification capability of MLLMs, RemoteZero reduces label dependence while providing a scalable route toward self-improving geospatial reasoning.

Overall, RemoteZero substantially outperforms general-purpose MLLMs and achieves a test Acc@0.5 of 71.29\% on EarthReason, surpassing the strongest supervised baseline by 3.18 percentage points without using labels for policy optimization. Beyond reasoning, RemoteZero extends effectively to visual grounding and referring segmentation, achieving competitive performance against specialized supervised methods. Its performance further improves consistently across self-evolution rounds and continues to benefit from an expanding pool of unlabeled training data. Together, these results bring us one step closer to Earth observation models capable of continual self-improvement.

The contributions of this work are summarized as follows:
\begin{itemize}
     \item We introduce RemoteZero, a label-free reinforcement learning framework that enables policy optimization directly on unlabeled Earth observation data.

    \item We explore a self-evolution paradigm for remote sensing MLLMs, offering a scalable route to continual improvement without additional supervision.

    \item We show that RemoteZero outperforms strong supervised baselines, transfers across spatial prediction tasks, and benefits from increasing training data.

\end{itemize}

\label{sec:intro}
\vspace{-1mm}

%% file: sections/motivation.tex
\section{Motivation}

Our pursuit of a zero-label approach to geospatial reasoning is based on the following two observations.

\subsection{Generation--Verification Asymmetry}
RemoteZero is first motivated by a general generation--verification asymmetry in MLLMs~\cite{he2026far}. 
For many structured prediction problems, generating a valid solution requires searching over a large output space, while verification only evaluates a specified candidate. 

Let $x$ denote an input and $z \in \mathcal{Z}$ denote a candidate solution. 
A generation model must produce a solution with high task utility:
\[
z^\star \in \arg\max_{z \in \mathcal{Z}} U(x,z),
\]
where $U(x,z)$ measures whether $z$ satisfies the task requirement. 
This formulation requires both evaluating candidate quality and searching over the output space $\mathcal{Z}$. 
For MLLMs, the difficulty is further amplified by autoregressive decoding: a structured output must be produced through a sequence of token decisions, and errors in intermediate tokens can invalidate the final solution.

Verification has a more constrained form:
\[
V(x,z) \rightarrow s \approx U(x,z),
\]
where the candidate $z$ is already specified and the model only estimates its compatibility with the input. 
Thus, verification removes the open-ended search component and reduces the output interface to a scalar or binary judgment. 

To examine whether this asymmetry holds in geospatial reasoning, we evaluate Qwen3-VL-8B on the EarthReason validation set under two settings. Directly generating a target region from an image--query pair yields only 43.88\% Acc@0.5. When given the ground-truth region and asked to judge its semantic consistency with the query, the model achieves 66.73\% verification accuracy. This contrast provides direct empirical evidence that region verification is substantially more reliable than coordinate generation.

\subsection{Overhead Imaging Prior in Remote Sensing}

Remote sensing imagery also provides a favorable domain for region-level verification. 
Most Earth observation images are captured from overhead or near-nadir viewpoints, where objects and land-cover areas are projected onto a nearly two-dimensional ground plane. 
Compared with perspective images, this view substantially reduces depth ambiguity, severe occlusion, and cross-object overlap.

Therefore, a cropped candidate region is less likely to contain multiple heavily overlapping objects from different depths. 
With moderate contextual padding, the crop is usually dominated by the intended facility, object, or functional area, while still preserving useful local cues such as nearby roads, buildings, fields, or open space. 
Thus, the semantic evidence inside a candidate crop can often be attributed to the proposed region itself rather than to unrelated overlapping content. These properties make candidate-region verification particularly suitable for geospatial reasoning.

%% file: sections/method.tex
\section{RemoteZero}
\label{sec:method}

\begin{figure*}[t]
  \centering
  \includegraphics[width=1.0\textwidth]{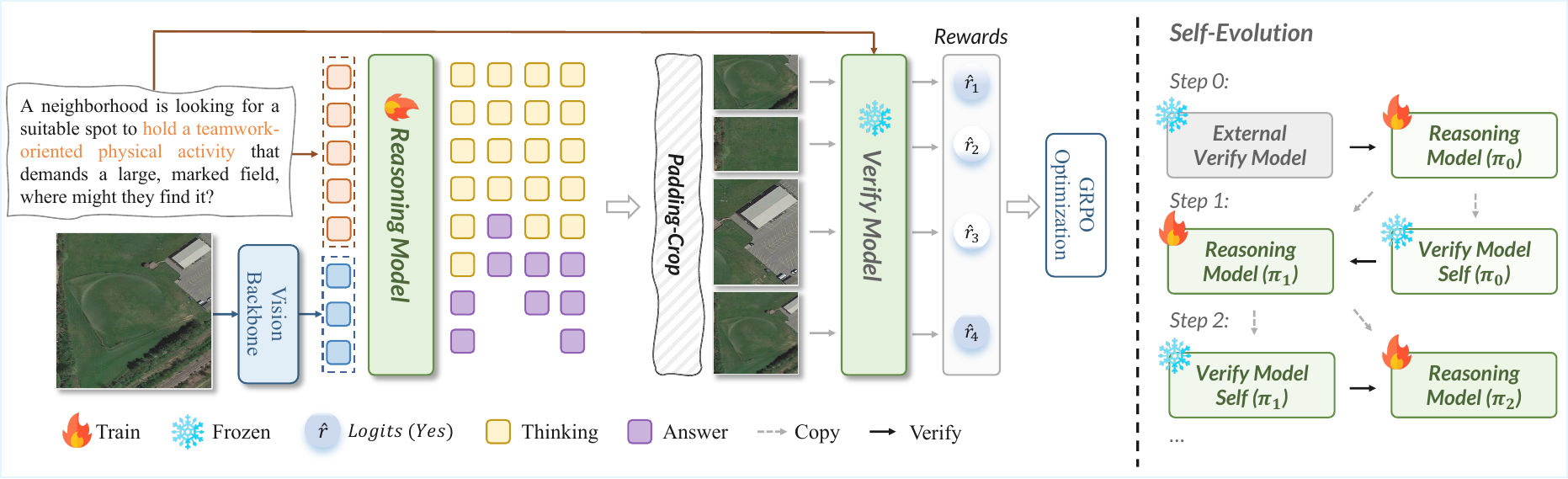}
  \caption{Overview of RemoteZero. The model generates a reasoning chain and a candidate box, which is converted into a padded crop and scored by a verifier for semantic consistency with the query. This score, combined with an area penalty, serves as the intrinsic reward for GRPO without ground-truth coordinates. RemoteZero further enables iterative self-evolution by reusing the frozen policy from the previous round as the verifier for the next round.}
  \label{overview}
\end{figure*}

\subsection{Problem Formulation}

Given a remote sensing image $\mathcal{I}$ and a query $\mathcal{Q}$, the policy model $\pi_\theta$ generates a reasoning chain $c$ and a spatial prediction $b$:
\begin{equation}
(c, b) \sim \pi_\theta(\cdot \mid \mathcal{I}, \mathcal{Q}).
\end{equation}
The goal is to learn a policy that predicts regions satisfying the query. 
We formulate this objective through a reward function $\mathcal{R}(b; \mathcal{I}, \mathcal{Q})$ defined over the predicted region.

Existing GRPO-based reinforcement learning methods compute the reward by comparing the predicted region $b$ with a human-annotated target $b_{gt}$:
\begin{equation}
\mathcal{R}_{ext}(b) = S_{\mathrm{gt}}(b, b_{gt}),
\end{equation}
where $S_{\mathrm{gt}}(\cdot)$ denotes a spatial matching score based on ground-truth annotations. 
Therefore, the reward can only be computed when annotated spatial labels are available.

RemoteZero replaces this annotation-dependent reward with semantic verification. 
Given the visual region $\mathcal{I}_b$ corresponding to the prediction $b$, the reward is defined as:
\begin{equation}
\mathcal{R}_{int}(b) = V(\mathcal{I}_b, \mathcal{Q}),
\end{equation}
where $V(\cdot)$ outputs a scalar score measuring whether the predicted region satisfies the query.

This reformulation changes the reward source from ground-truth spatial matching to query-region semantic verification, enabling policy learning without labels.

\subsection{Training}
Follow the previous work~\cite{yao2026remotereasoner,yao2026remoteagent}, we utilize GRPO to fine-tune the policy model. 
As shown in Fig.~\ref{overview}, Given a remote sensing image $\mathcal{I}$ and an implicit user query $\mathcal{Q}$, the policy model $\pi_\theta$ (an MLLM) generates a sequence consisting of a reasoning chain $\mathbf{c}$ and a predicted bounding box $\mathbf{b} = [x_1, y_1, x_2, y_2]$. This process is formulated as sampling from the policy:

\begin{equation}
    \mathbf{c}, \mathbf{b} \sim \pi_\theta(\cdot | \mathcal{I}, \mathcal{Q})
\end{equation}

Unlike standard supervised approaches, this generation process is not constrained by ground-truth coordinates but is guided solely by the subsequent reward signal.

To evaluate the correctness of the predicted location $\mathbf{b}$, we isolate the region of interest. We define a deterministic cropping function $\mathcal{T}(\cdot)$ that extracts the image patch corresponding to $\mathbf{b}$. To preserve local context essential for verification (e.g., surrounding roads or terrain), we apply a relaxed margin ratio $\alpha$ to the bounding box before cropping:

\begin{equation}
    \mathcal{I}_{crop} = \mathcal{T}(\mathcal{I}, \mathbf{b}, \alpha).
\end{equation}

The cropped patch $\mathcal{I}_{crop}$ is fed into a verifier model $V$. The verifier assesses the semantic entailment between the visual crop and the original query $\mathcal{Q}$. It outputs a scalar confidence score $s \in [0, 1]$, representing the probability that the cropped region semantically satisfies the query condition:

\begin{equation}
    s = V(\mathcal{I}_{crop}, \mathcal{Q}).
\end{equation}

It is worth noting that this formulation is agnostic to the specific instantiation of $V$. Whether $V$ is an external superior model or the policy $\pi_\theta$ itself, the mathematical interface remains unified.

The final reward $r$ driving the GRPO optimization is a composite of the verification confidence and a regularization term. Since the verifier might trivially award high scores to overly large crops (which are more likely to contain the target but lack precision), we introduce an area penalty:
\begin{equation}
    r(\mathbf{b}, \mathcal{Q}) = s - \lambda \cdot \max\left(0, \frac{\text{Area}(\mathbf{b})}{\text{Area}(\mathcal{I})} - \tau\right),
\end{equation}
where $\lambda$ controls the penalty strength and $\tau$ is a threshold for acceptable area proportion. This reward is then used to compute the advantages for the group of sampled outputs in the GRPO objective.

\subsection{Self-Evolution}

The training procedure above can use a stronger model as the verifier $V$, but such an external verifier is not essential. Since verification is more reliable than generation in this setting, a model's own verification capability can also provide useful feedback for improving its future predictions. This observation motivates a self-evolution procedure in which the policy obtained from one round serves as the verifier for the next round.


We first obtain an initial policy $\pi_{\theta}^{(0)}$ using an external verifier under the training procedure described in Sec.~3.2. Self-evolution then proceeds for $K$ rounds. At round $k>0$, a trainable policy $\pi_{\theta}^{(k)}$ is initialized from $\pi_{\theta}^{(k-1)}$, while a separate copy of the previous policy is frozen as the verifier for the current round:
\begin{equation}
V^{(k)}(\cdot) \leftarrow \pi_{\theta}^{(k-1)}(\cdot).
\end{equation}

For each candidate box $b$ generated by $\pi_{\theta}^{(k)}$, the round-specific reward follows the same crop-based verification and area regularization introduced in Sec.~3.2:
\begin{equation}
\begin{aligned}
r^{(k)}(b,Q)
= {} & V^{(k)}\!\left(\mathcal{T}(I,b,\alpha),Q\right) \\
& - \lambda \max\left(0,
\frac{\operatorname{Area}(b)}{\operatorname{Area}(I)}-\tau\right).
\end{aligned}
\end{equation}
During optimization, $V^{(k)}$ remains fixed, while $\pi_{\theta}^{(k)}$ is updated with GRPO. Once the round is completed, the updated policy is carried forward and frozen as the verifier for round $k+1$. Therefore, the external verifier is needed only to initialize the process. Subsequent rounds are driven by the preceding policy's verification feedback and introduce no coordinate annotations.

%% file: sections/experiments.tex
\vspace{-2mm}
\section{Experimental Results}

\subsection{Experimental Setup}

\subsubsection{Experimental settings}
We instantiate RemoteZero with Qwen3-VL-8B-Instruct~\cite{bai2025qwen3vltechnicalreport} and train it with GRPO~\cite{shao2024deepseekmath} using LoRA fine-tuning~\cite{hu2022lora}. Training is conducted on 8 $\times$ GPUs with DeepSpeed ZeRO-2~\cite{rasley2020deepspeed} in bfloat16. The model is optimized for 10 epochs with a learning rate of $5\times10^{-6}$, using a per-device batch size of 6 and gradient accumulation of 8 steps. For each prompt, GRPO samples 4 generations with temperature 0.9. The maximum sequence length is set to 2048, and the maximum image size is capped at 802{,}816 pixels. Unless otherwise stated, all experiments use the same training configuration.

\subsubsection{Datasets and Baselines}
We conduct the geospatial reasoning experiments on EarthReason~\cite{li2025segearth}. 
To evaluate the applicability of RemoteZero to other EO tasks, we use DIOR-RSVG~\cite{zhan2023rsvg} for grounding and RRSIS-D~\cite{yuan2024rrsis} for referring segmentation. We select Qwen2.5-VL-7B~\cite{bai2025qwen2}, DeepSeek-VL2~\cite{wu2024deepseek}, InternVL3.5~\cite{wang2025internvl3}, VLM-R1~\cite{shen2025vlm}, GeoChat~\cite{kuckreja2024geochat}, and RemoteReasoner~\cite{yao2026remotereasoner} as baselines. 
For visual grounding, the baselines include SkyEyeGPT~\cite{zhan2025skyeyegpt}, GeoChat~\cite{kuckreja2024geochat}, LHRS-Bot~\cite{muhtar2024lhrs}, RemoteSAM~\cite{yao2025remotesam}, EarthDial~\cite{soni2025earthdial}, EarthGPT~\cite{zhang2024earthgpt}, and GeoZero~\cite{wang2025geozero}. 
For referring segmentation, we compare with LISA~\cite{lai2024lisa}, PixelLM~\cite{ren2024pixellm}, Text4Seg++~\cite{lan2025text4seg++}, GeoGround~\cite{zhou2024geoground}, SegEarth-R1~\cite{li2025segearth}, and ProVG~\cite{li2026provg}.

\subsection{Main Results}

Table~\ref{tab:main} compares RemoteZero with general-purpose MLLMs, remote-sensing MLLMs, and supervised geospatial reasoning methods on EarthReason. 
Existing MLLMs exhibit limited performance on implicit geospatial localization. 
Among the general-purpose models, Qwen3-VL-8B achieves the strongest test Acc@0.5 of 48.10\%, while reasoning-oriented VLM-R1 reaches only 33.31\%. 
GeoChat also performs poorly despite its remote-sensing specialization, suggesting that domain-specific visual-language alignment alone is insufficient for resolving complex user intent into precise regions.

RemoteZero substantially improves over these baselines. 
With an external verifier, RemoteZero* achieves 65.05\% test Acc@0.5 without using human-provided solution labels for policy optimization, approaching the fully supervised RemoteReasoner. 
After self-evolution, RemoteZero further improves to 71.29\%, surpassing RemoteReasoner by 3.18 percentage points and achieving the best Acc@0.5 on both splits. 
Nevertheless, RemoteZero obtains a lower test gIoU than RemoteReasoner, which indicates that the current verifier reward is more effective at identifying semantically correct regions than at calibrating precise spatial extents. We view this as an important direction for future improvement.

\begin{table}[t]
\centering
\caption{Comparison with other MLLMs on EarthReason. `*' denotes supervised by an external verifier.}
\setlength{\tabcolsep}{7pt}
\begin{tabular}{l|cc|cc}
\toprule
\multirow{2}{*}{Method} & \multicolumn{2}{c|}{Acc@0.5} & \multicolumn{2}{c}{gIoU} \\
\cline{2-5}
                        & Val   & Test  & Val   & Test  \\
\hline
Qwen2.5-VL-7B                  & 41.21 & 45.82 & 38.77 & 41.80 \\
Qwen3-VL-8B                    & 43.88 & 48.10 & 41.03 & 44.69 \\
DeepSeek-VL2                    & 12.08 & 12.67 & 17.51 & 18.62 \\
InternVL3.5                      & 4.88  & 5.26  & 5.83  & 6.52  \\
VLM-R1                          & 34.64 & 33.31 & 29.67 & 29.44 \\
GeoChat                          & 10.10 & 8.89  & 12.57 & 11.44 \\
RemoteReasoner                 & \underline{66.51} & \underline{68.11} & \textbf{67.04} & \textbf{69.29} \\
\hline
\textbf{RemoteZero*} & 65.38 & 65.05 & 58.10 & 57.95 \\ 
\textbf{RemoteZero} & \textbf{69.96} & \textbf{71.29} & \underline{61.54} & \underline{61.70} \\
\bottomrule
\end{tabular}
\label{tab:main}
\end{table}

\begin{table}[t]
\centering
\caption{Ablation study of reward functions.}
\setlength{\tabcolsep}{15pt}
\begin{tabular}{cc|cc}
\hline
\multirow{2}{*}{Verify} & \multirow{2}{*}{Area} & \multicolumn{2}{c}{Acc@0.5} \\
\cline{3-4}
 & & Val & Test \\
\hline
$\checkmark$ &  & 65.20 & 65.88 \\
$\checkmark$ & $\checkmark$ & \textbf{69.96} & \textbf{71.29} \\
\hline
\end{tabular}

\label{tab:reward}
\end{table}

\begin{table}[t]
\centering
\setlength{\tabcolsep}{9pt}
\caption{Ablation study of cropping strategies with different context padding ratios.}
\begin{tabular}{l|c|cc}
\toprule
\multirow{2}{*}{Crop Strategy}
& \multirow{2}{*}{Padding}
& \multicolumn{2}{c}{Acc@0.5} \\
\cline{3-4}
& & Val & Test \\
\hline
Strict  & --    & 64.61 & 65.13 \\
\hline
\multirow{3}{*}{Context}
        & 10\%  & 66.22 & 68.84 \\
        & 15\%  & \textbf{69.96} & \textbf{71.29} \\
        & 20\%  & 67.06 & 68.17 \\
\bottomrule
\end{tabular}
\label{tab:crop}
\end{table}

\begin{table}[t]
\centering
\caption{Ablation study of different verifier models.}
\renewcommand{\arraystretch}{1.02}
\setlength{\tabcolsep}{6.2pt}
\begin{tabular}{l|cc|c}
\toprule
\multirow{2}{*}{Verify Model}
& \multicolumn{2}{c|}{Acc@0.5} & \multirow{2}{*}{Verify Score }\\
\cline{2-3}
 & Val & Test & \\
\hline
Qwen2.5-VL-72B & \textbf{65.38} & \textbf{65.05} & 66.19 \\
Qwen2.5-VL-32B & 63.01 & 62.85 & 63.25 \\
Qwen3-VL-32B & 64.55 & 64.90 & \textbf{67.30} \\
Qwen3-VL-8B & 63.52 & 64.21 & \underline{66.73} \\
RemoteReasoner & 63.68 & 64.17 & 65.48 \\
\bottomrule
\end{tabular}
\label{tab:verify_model}
\end{table}

\subsection{Ablation Studies}

\subsubsection{Ablation on Reward Design}

To assess whether semantic verification alone is sufficient for spatially precise reward learning, We compare the verifier-only reward with the full reward that further includes area penalty, as reported in Table~\ref{tab:reward}. 
Using only the verification score already provides a meaningful training signal, achieving 65.20\% validation and 65.88\% test Acc@0.5. 
However, this reward can favor overly large boxes, since larger regions are more likely to contain query-relevant content and receive positive verifier feedback. 
Adding the area penalty improves performance to 69.96\% on validation and 71.29\% on test, indicating that a lightweight geometric constraint is necessary to prevent trivial large-region predictions. 
These results suggest that semantic verification provides the core learning signal, while area penalty further improves localization precision.
\subsubsection{Ablation on Cropping Strategy}

We compare strict cropping with context crops using different padding ratios to examine how the amount of visual context affects verification-based training. 
As illustrated in Table~\ref{tab:crop}, strict cropping achieves 64.61\% validation and 65.13\% test Acc@0.5, suggesting that the predicted box alone may discard surrounding cues needed for interpreting functional geospatial intent. 
Introducing contextual padding improves performance, with 15\% padding reaching the best results of 69.96\% on validation and 71.29\% on test. 
However, further increasing the padding to 20\% leads to lower accuracy, indicating that excessive context may introduce irrelevant regions and weaken the verifier's region-specific judgment.

\subsubsection{Ablation on Verifier Models}

To study the effect of verifier choice, we train the same base model using rewards provided by different verifier models, as shown in Table~\ref{tab:verify_model}. 
The verification score measures the average probability of answering \texttt{Yes} when the verifier is given a ground-truth crop together with its query, while Val and Test report the downstream Acc@0.5 of the trained base model. 
Within the same model family, larger verifiers generally achieve higher verification scores, and the newer Qwen3-VL models also exhibit stronger verification capability. 
However, higher verification scores do not always lead to proportional gains in downstream localization accuracy: Qwen2.5-VL-72B achieves the best Val and Test results, whereas Qwen3-VL-32B obtains the highest verification score.  
Considering the substantial training cost of larger verifiers and the relatively small difference in final accuracy, we adopt Qwen3-VL-8B as the default verifier, which provides a favorable trade-off between verification quality and training efficiency.

\subsection{Further Analysis}

\subsubsection{Is the Verification Reward Vulnerable to Trivial Coverage Bias?}

To examine whether the verification reward reflects genuine query-region compatibility, we evaluate the verifier under different visual inputs, as reported in Table~\ref{tab:reward_hacking}. 
The verifier assigns high confidence to ground-truth regions, while shifted boxes and target-free regions receive substantially lower rewards, indicating that the reward is sensitive to spatial relevance rather than arbitrary visual content. 
The padded crop achieves the highest verification confidence and final accuracy, suggesting that moderate local context helps the verifier interpret functional geospatial intent. 
In contrast, the whole image receives a much lower reward despite containing the target and more visual content, showing that the verifier does not simply favor larger regions. 
These results indicate that the verification reward primarily captures query-region compatibility, while padded crop provides an effective balance between semantic context and spatial specificity.

\begin{table}[t]
\centering
\caption{Analysis of verification reward behavior under different visual inputs. 
}
\setlength{\tabcolsep}{4.8pt}
\begin{tabular}{l|ccc}
\toprule
Input Type & Prob. (Yes) \% & Rewards & Acc@0.5 \\
\hline
GT box & 86.3 & 0.770 & 65.13 \\
Shifted box & 34.9 & 0.302 & 20.08 \\
\textbf{Padding image} & \textbf{93.5} & \textbf{0.896} & \textbf{71.29} \\
Whole image & 11.7 & 0.091 & 16.53 \\
No targets & 9.0 & 0.074 & 15.40 \\
\bottomrule
\end{tabular}

\label{tab:reward_hacking}
\end{table}

\subsubsection{Can RemoteZero Be Applied to Other Tasks?}

\begin{table}[t]
\centering

\caption{Application of RemoteZero to visual grounding and referring segmentation. 
}
\setlength{\tabcolsep}{8pt}
\begin{tabular}{@{}c|l|cc@{}}
\toprule
Task & Method & $Pr@0.5$ & $mIoU$ \\
\hline

\multirow{10}{*}{\shortstack[c]{Visual\\Grounding}}
& SkyEyeGPT   & 70.5 & -- \\
& GeoChat     & 62.0 & -- \\
& LHRS-Bot    & 73.5 & -- \\
& RemoteSAM   & 74.4 & 65.1 \\
& EarthDial   & 46.1 & 39.5 \\
& EarthGPT    & \textbf{76.7} & \textbf{69.3} \\
& RSGround-R1 & 71.8 & 63.4 \\
& TinyRS-R1   & 74.9 & -- \\
& GeoZero     & 75.7 & -- \\
& RemoteZero  & \underline{76.0} & \underline{68.9} \\

\hline

\multirow{7}{*}{\shortstack[c]{Referring\\Segmentation}}
& LISA        & -- & 26.8 \\
& PixelLM     & -- & 31.7 \\
& Tex4Seg++   & 73.2 & 62.8 \\
& GeoGround   & -- & 60.5 \\
& SegEarth-R1 & 67.6 & \textbf{66.4} \\
& ProVG       & \textbf{76.1} & \underline{65.4} \\
& RemoteZero$^\dagger$
               & \underline{75.3} & \underline{65.4} \\
\bottomrule
\end{tabular}

\label{tab:generalization}
\end{table}

To examine whether RemoteZero can be applied to other remote sensing tasks, we instantiate the proposed label-free training framework on visual grounding and further extend it to referring segmentation, as reported in Table~\ref{tab:generalization}. 
For visual grounding, RemoteZero achieves 76.0 Pr@0.5 and 68.9 mIoU, outperforming most specialized grounding methods and remaining close to the strongest supervised baseline. 
For referring segmentation, we use the bounding boxes predicted by the label-free RemoteZero grounding model as prompts for SAM~2, without introducing segmentation labels or segmentation-specific training for RemoteZero. 
This simple combination achieves 75.3 Pr@0.5 and 65.4 mIoU, reaching performance comparable to dedicated referring segmentation methods. 
These results show that RemoteZero is not tied to a particular geospatial reasoning task: its verification-based reward formulation can be instantiated for other spatial prediction tasks and can also provide effective intermediate outputs for downstream pixel-level applications.

\begin{figure}[t]
    \centering
    \includegraphics[width=0.5\linewidth]{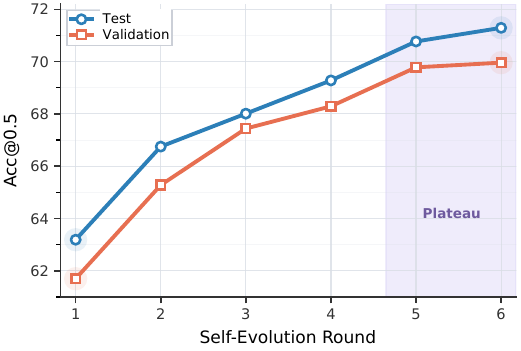}
    \caption{
    Performance across six self-evolution rounds on EarthReason.
    Both validation and test Acc@0.5 improve consistently, while the gains diminish in later rounds, indicating saturation under a fixed training set.
    }
    \label{fig:self_evolution_rounds}
\end{figure}

\begin{figure}[t]
    \centering
    \includegraphics[width=0.5\linewidth]{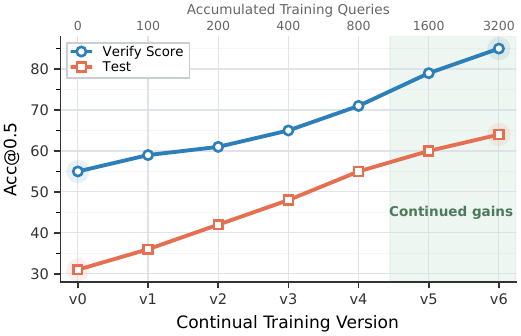}
    \caption{Performance under continual data scaling on EarthReason. 
    The accumulated training queries double at each version. 
    Both the verifier's Verify Score and the policy's test Acc@0.5 continue to improve, suggesting that the earlier plateau is mainly associated with limited data diversity.}
    \label{fig:continual_evolution}
\end{figure}

\begin{figure*}[t]
  \centering
  \includegraphics[width=1\textwidth]{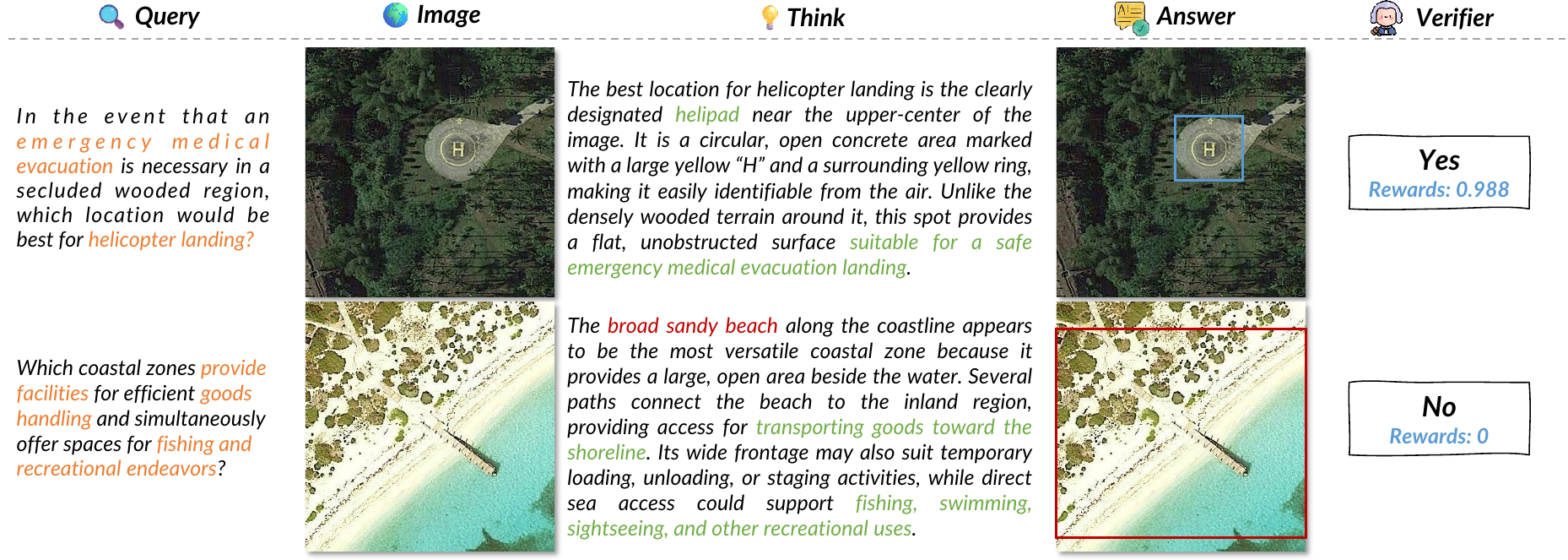}
  \caption{Qualitative case studies. The verifier rewards accurate localization (top) while rejecting a semantically mismatched, overly large prediction (bottom).}
  \label{vis}
\end{figure*}

\subsubsection{How Many Self-Evolution Rounds Are Needed?}

We also evaluate RemoteZero after each evolution round on EarthReason to study how iterative self-evolution affects performance. 
As shown in Fig.~\ref{fig:self_evolution_rounds}, both validation and test accuracy improve consistently across the six rounds. 
From the first to the sixth round, validation Acc@0.5 increases from 61.70\% to 69.96\%, while test accuracy improves from 63.19\% to 71.29\%, corresponding to gains of 8.26 and 8.10 percentage points, respectively. 
The largest improvements occur during the early rounds, whereas the gains from the fifth to the sixth round decrease to 0.18 and 0.52 percentage points on the validation and test sets. 
This trend suggests that self-evolution is approaching saturation by the sixth round. 
We therefore use the 6th-round model for the final evaluation, providing a practical balance between performance and training cost.

\subsubsection{Can Self-Evolution Continue with Expanding Data?}

The previous experiment shows that self-evolution gradually saturates after six rounds when the training data remain fixed. 
To examine whether this plateau results from limited data scale, we simulate a continually expanding data pool by sampling queries from EarthReason and doubling the accumulated training data at each version. 
As shown in Fig.~\ref{fig:continual_evolution}, unlike the saturation observed with a fixed training set, both metrics retain substantial gains in later versions. 
These results suggest that the earlier plateau is primarily associated with the finite data pool rather than an inherent limit of self-evolution. 
As increasingly diverse queries become available, the verifier becomes more reliable and provides stronger feedback for subsequent policy optimization.

\subsubsection{Qualitative Case Study}
Fig.~\ref{vis} presents representative successful and failed cases of RemoteZero. In the first row, the model correctly interprets the emergency evacuation intent, identifies the marked helipad as a flat and unobstructed landing area, and produces a compact bounding box around it. The verifier confirms that the predicted region satisfies the query, yielding a high reward of 0.988. In the second row, however, the model overemphasizes the broad and open characteristics of the sandy beach while overlooking the query's requirement for human-made facilities that support efficient goods handling. Consequently, it incorrectly reasons that the beach is the target and predicts an excessively large region instead of focusing on the relevant coastal infrastructure. The verifier successfully detects this semantic mismatch, outputs \textit{No}, and assigns zero reward. This failure case demonstrates that verification-based feedback can reject superficially plausible predictions when their localized regions do not satisfy all essential constraints expressed in the query.

%% file: sections/relatedwork.tex
\section{Related Work}
\label{sec:related_work}

\subsection{Remote Sensing Multi-modal Models}
The adaptation of MLLMs to remote sensing~\cite{GeoMag,ma2026weaveearth,min2026remoteshield,ye2026crossearthsar,ye2026objectfidelity} began with domain-specific captioning and grounded dialogue, represented by RSGPT~\cite{hu2025rsgpt} and GeoChat~\cite{kuckreja2024geochat}. 
EarthGPT~\cite{zhang2024earthgpt} and SkyEyeGPT~\cite{zhan2025skyeyegpt} then expanded toward unified multi-sensor interpretation and instruction tuning. 
Subsequent works improved granularity and interaction through relational reasoning in EarthVQA~\cite{wang2024earthvqa}, VGI-enhanced training in LHRS-Bot~\cite{muhtar2024lhrs}, fine-grained instruction tuning in SkySenseGPT~\cite{luo2024sky}, and visual prompting in EarthMarker~\cite{zhang2024earthmarker}. 
TEOChat~\cite{irvin2024teochat} and RingMoGPT~\cite{wang2024ringmogpt} further extended temporal and grounded understanding, while RSUniVLM~\cite{liu2024rsunivlm}, Falcon~\cite{yao2025falcon}, and EagleVision~\cite{jiang2025eaglevisionobjectlevelattributemultimodal} moved toward unified pixel-level and object-centric interpretation.

\subsection{Geospatial Reasoning Models}
Remote sensing research has recently moved beyond standard perception~\cite{liu2024remoteclip,huang2026rsgroundr1} toward complex geospatial reasoning with MLLMs. SegEarth-R1/2~\cite{li2025segearth,xin2026segearth} addressed implicit queries through pixel-level reasoning, followed by RemoteReasoner~\cite{yao2026remotereasoner}, which introduced a unified RL workflow for autonomous multi-granularity analysis. Subsequent work increasingly adopted RFT and CoT: Geo-R1~\cite{zhang2025geor1} applied RFT to few-shot referring expression understanding, GeoVLM-R1~\cite{fiaz2025geovlm} designed task-aware rewards for diverse observation tasks, and RSThinker~\cite{liu2025towards} developed perceptually grounded CoT to produce verifiable outputs. More recently, GeoZero~\cite{wang2025geozero} encouraged emergent reasoning without predefined CoT supervision through answer-anchored policy optimization, while GeoReason~\cite{li2026georeason} used consistency-aware RL to better align internal reasoning with final predictions.

%% file: sections/conclusion.tex
\section{Conclusion}
\label{sec:conclusion}

We presented RemoteZero, a label-free framework for reinforcement-based geospatial reasoning. 
RemoteZero replaces human-provided solution labels with intrinsic semantic verification and further supports self-evolution by reusing previous-round models as verifiers. Experiments on EarthReason show that RemoteZero achieves 71.29\% Acc@0.5, surpassing the strongest supervised baseline by 3.18 percentage points. 
We are excited by the broader possibility that our method may offer a general route for MLLMs to learn from unlabeled data and gain continually improvement.

%% file: main.bib
@inproceedings{xin2026segearth,
  title={Segearth-r2: Towards comprehensive language-guided segmentation for remote sensing images},
  author={Xin, Zepeng and Li, Kaiyu and Chen, Luodi and Li, Wanchen and Yuchen, Xiao and Qiao, Hui and Zhang, Weizhan and Meng, Deyu and Cao, Xiangyong},
  booktitle={Proceedings of the IEEE/CVF Conference on Computer Vision and Pattern Recognition},
  pages={13199--13210},
  year={2026}
}

@article{li2025segearth,
  title={Segearth-r1: Geospatial pixel reasoning via large language model},
  author={Li, Kaiyu and Xin, Zepeng and Pang, Li and Pang, Chao and Deng, Yupeng and Yao, Jing and Xia, Guisong and Meng, Deyu and Wang, Zhi and Cao, Xiangyong},
  journal={arXiv preprint arXiv:2504.09644},
  year={2025}
}

@article{hu2025rsgpt,
  title={Rsgpt: A remote sensing vision language model and benchmark},
  author={Hu, Yuan and Yuan, Jianlong and Wen, Congcong and Lu, Xiaonan and Liu, Yu and Li, Xiang},
  journal={ISPRS Journal of Photogrammetry and Remote Sensing},
  volume={224},
  pages={272--286},
  year={2025},
  publisher={Elsevier}
}

@inproceedings{wang2024earthvqa,
  title={Earthvqa: Towards queryable earth via relational reasoning-based remote sensing visual question answering},
  author={Wang, Junjue and Zheng, Zhuo and Chen, Zihang and Ma, Ailong and Zhong, Yanfei},
  booktitle={Proceedings of the AAAI conference on artificial intelligence},
  volume={38},
  number={6},
  pages={5481--5489},
  year={2024}
}

@article{zhang2024earthgpt,
  title={EarthGPT: A universal multimodal large language model for multisensor image comprehension in remote sensing domain},
  author={Zhang, Wei and Cai, Miaoxin and Zhang, Tong and Zhuang, Yin and Mao, Xuerui},
  journal={IEEE Transactions on Geoscience and Remote Sensing},
  volume={62},
  pages={1--20},
  year={2024},
  publisher={IEEE}
}

@article{zhang2024earthmarker,
  title={Earthmarker: A visual prompting multi-modal large language model for remote sensing},
  author={Zhang, Wei and Cai, Miaoxin and Zhang, Tong and Zhuang, Yin and Li, Jun and Mao, Xuerui},
  journal={IEEE Transactions on Geoscience and Remote Sensing},
  year={2024},
  publisher={IEEE}
}

@article{wang2024ringmogpt,
  title={Ringmogpt: A unified remote sensing foundation model for vision, language, and grounded tasks},
  author={Wang, Peijin and Hu, Huiyang and Tong, Boyuan and Zhang, Ziqi and Yao, Fanglong and Feng, Yingchao and Zhu, Zining and Chang, Hao and Diao, Wenhui and Ye, Qixiang and others},
  journal={IEEE Transactions on Geoscience and Remote Sensing},
  year={2024},
  publisher={IEEE}
}

@article{liu2024rsunivlm,
  title={Rsunivlm: A unified vision language model for remote sensing via granularity-oriented mixture of experts},
  author={Liu, Xu and Lian, Zhouhui},
  journal={arXiv preprint arXiv:2412.05679},
  year={2024}
}

@article{zhan2025skyeyegpt,
  title={Skyeyegpt: Unifying remote sensing vision-language tasks via instruction tuning with large language model},
  author={Zhan, Yang and Xiong, Zhitong and Yuan, Yuan},
  journal={ISPRS Journal of Photogrammetry and Remote Sensing},
  volume={221},
  pages={64--77},
  year={2025},
  publisher={Elsevier}
}

@article{luo2024sky,
    title={SkySenseGPT: A Fine-Grained Instruction Tuning Dataset and Model for Remote Sensing Vision-Language Understanding},
    author={Luo, Junwei and Pang, Zhen and Zhang, Yongjun and Wang, Tingzhu and Wang, Linlin and Dang, Bo and Lao, Jiangwei and Wang, Jian and Chen, Jingdong and Tan, Yihua and Li, Yansheng},
    journal={arXiv preprint arXiv:2406.10100},
    year={2024}
}

@article{yao2025falcon,
  title={Falcon: A Remote Sensing Vision-Language Foundation Model},
  author={kelu, Yao and Nuo, Xu and Rong, Yang and Yingying, Xu and Zhuoyan, Gao and Titinunt, Kitrungrotsakul and yi, Ren and Pu, Zhang and Jin, Wang and Ning, Wei and Chao, Li},
  journal={arXiv preprint arXiv:2503.11070},
  year={2025}
}

@inproceedings{irvin2024teochat,
  title={TEOChat: A Large Vision-Language Assistant for Temporal Earth Observation Data},
  author={Irvin, Jeremy Andrew and Liu, Emily Ruoyu and Chen, Joyce Chuyi and Dormoy, Ines and Kim, Jinyoung and Khanna, Samar and Zheng, Zhuo and Ermon, Stefano},
  booktitle={International Conference on Learning Representations},
  year={2025}
}

@article{yao2025remotesam,
  title={RemoteSAM: Towards Segment Anything for Earth Observation},
  author={Yao, Liang and Liu, Fan and Chen, Delong and Zhang, Chuanyi and Wang, Yijun and Chen, Ziyun and Xu, Wei and Di, Shimin and Zheng, Yuhui},
  journal={arXiv preprint arXiv:2505.18022},
  year={2025}
}

@misc{ye2026objectfidelity,
      title={Object Fidelity Diffusion for Remote Sensing Image Generation}, 
      author={Ziqi Ye and Shuran Ma and Jie Yang and Xiaoyi Yang and Yi Yang and Ziyang Gong and Xue Yang and Haipeng Wang},
      year={2026},
      eprint={2508.10801},
      archivePrefix={arXiv},
      primaryClass={cs.CV},
      url={https://arxiv.org/abs/2508.10801}, 
}

@misc{ye2026crossearthsar,
      title={CrossEarth-SAR: A SAR-Centric and Billion-Scale Geospatial Foundation Model for Domain Generalizable Semantic Segmentation}, 
      author={Ziqi Ye and Ziyang Gong and Ning Liao and Xiaoxing Hu and Di Wang and Hongruixuan Chen and Chen Huang and Yiguo He and Yuru Jia and Xiaoxing Wang and Haipeng Wang and Xue Yang and Junchi Yan},
      year={2026},
      eprint={2603.12008},
      archivePrefix={arXiv},
      primaryClass={cs.CV},
      url={https://arxiv.org/abs/2603.12008}, 
}

@inproceedings{muhtar2024lhrs,
  title={Lhrs-bot: Empowering remote sensing with vgi-enhanced large multimodal language model},
  author={Muhtar, Dilxat and Li, Zhenshi and Gu, Feng and Zhang, Xueliang and Xiao, Pengfeng},
  booktitle={European Conference on Computer Vision},
  pages={440--457},
  year={2024},
  organization={Springer}
}

@article{zhou2024geoground,
  title={Geoground: A unified large vision-language model. for remote sensing visual grounding},
  author={Zhou, Yue and Lan, Mengcheng and Li, Xiang and Ke, Yiping and Jiang, Xue and Feng, Litong and Zhang, Wayne},
  journal={arXiv preprint arXiv:2411.11904},
  year={2024}
}

@article{xu2025towards,
  title={Towards large reasoning models: A survey of reinforced reasoning with large language models},
  author={Xu, Fengli and Hao, Qianyue and Zong, Zefang and Wang, Jingwei and Zhang, Yunke and Wang, Jingyi and Lan, Xiaochong and Gong, Jiahui and Ouyang, Tianjian and Meng, Fanjin and others},
  journal={arXiv preprint arXiv:2501.09686},
  year={2025}
}

@inproceedings{lai2024lisa,
  title={Lisa: Reasoning segmentation via large language model},
  author={Lai, Xin and Tian, Zhuotao and Chen, Yukang and Li, Yanwei and Yuan, Yuhui and Liu, Shu and Jia, Jiaya},
  booktitle={Proceedings of the IEEE/CVF conference on computer vision and pattern recognition},
  pages={9579--9589},
  year={2024}
}

@inproceedings{ren2024pixellm,
  title={Pixellm: Pixel reasoning with large multimodal model},
  author={Ren, Zhongwei and Huang, Zhicheng and Wei, Yunchao and Zhao, Yao and Fu, Dongmei and Feng, Jiashi and Jin, Xiaojie},
  booktitle={Proceedings of the IEEE/CVF conference on computer vision and pattern recognition},
  pages={26374--26383},
  year={2024}
}

@article{liu2024remoteclip,
  title={Remoteclip: A vision language foundation model for remote sensing},
  author={Liu, Fan and Chen, Delong and Guan, Zhangqingyun and Zhou, Xiaocong and Zhu, Jiale and Ye, Qiaolin and Fu, Liyong and Zhou, Jun},
  journal={IEEE Transactions on Geoscience and Remote Sensing},
  volume={62},
  pages={1--16},
  year={2024},
  publisher={IEEE}
}

@inproceedings{rasley2020deepspeed,
  title={Deepspeed: System optimizations enable training deep learning models with over 100 billion parameters},
  author={Rasley, Jeff and Rajbhandari, Samyam and Ruwase, Olatunji and He, Yuxiong},
  booktitle={Proceedings of the 26th ACM SIGKDD international conference on knowledge discovery \& data mining},
  pages={3505--3506},
  year={2020}
}

@article{hu2022lora,
  title={Lora: Low-rank adaptation of large language models.},
  author={Hu, Edward J and Shen, Yelong and Wallis, Phillip and Allen-Zhu, Zeyuan and Li, Yuanzhi and Wang, Shean and Wang, Lu and Chen, Weizhu and others},
  journal={ICLR},
  volume={1},
  number={2},
  pages={3},
  year={2022}
}

@article{wu2024deepseek,
  title={Deepseek-vl2: Mixture-of-experts vision-language models for advanced multimodal understanding},
  author={Wu, Zhiyu and Chen, Xiaokang and Pan, Zizheng and Liu, Xingchao and Liu, Wen and Dai, Damai and Gao, Huazuo and Ma, Yiyang and Wu, Chengyue and Wang, Bingxuan and others},
  journal={arXiv preprint arXiv:2412.10302},
  year={2024}
}

@article{shen2025vlm,
  title={Vlm-r1: A stable and generalizable r1-style large vision-language model},
  author={Shen, Haozhan and Liu, Peng and Li, Jingcheng and Fang, Chunxin and Ma, Yibo and Liao, Jiajia and Shen, Qiaoli and Zhang, Zilun and Zhao, Kangjia and Zhang, Qianqian and others},
  journal={arXiv preprint arXiv:2504.07615},
  year={2025}
}

@article{shao2024deepseekmath,
  title={Deepseekmath: Pushing the limits of mathematical reasoning in open language models},
  author={Shao, Zhihong and Wang, Peiyi and Zhu, Qihao and Xu, Runxin and Song, Junxiao and Bi, Xiao and Zhang, Haowei and Zhang, Mingchuan and Li, YK and Wu, Yang and others},
  journal={arXiv preprint arXiv:2402.03300},
  year={2024}
}

@article{bai2025qwen2,
  title={Qwen2. 5-vl technical report},
  author={Bai, Shuai and Chen, Keqin and Liu, Xuejing and Wang, Jialin and Ge, Wenbin and Song, Sibo and Dang, Kai and Wang, Peng and Wang, Shijie and Tang, Jun and others},
  journal={arXiv preprint arXiv:2502.13923},
  year={2025}
}

@inproceedings{kuckreja2024geochat,
  title={Geochat: Grounded large vision-language model for remote sensing},
  author={Kuckreja, Kartik and Danish, Muhammad Sohail and Naseer, Muzammal and Das, Abhijit and Khan, Salman and Khan, Fahad Shahbaz},
  booktitle={Proceedings of the IEEE/CVF conference on computer vision and pattern recognition},
  pages={27831--27840},
  year={2024}
}

@misc{huang2026rsgroundr1,
      title={RSGround-R1: Rethinking Remote Sensing Visual Grounding through Spatial Reasoning}, 
      author={Shiqi Huang and Shuting He and Bihan Wen},
      year={2026},
      eprint={2601.21634},
      archivePrefix={arXiv},
      primaryClass={cs.CV},
      url={https://arxiv.org/abs/2601.21634}, 
}

@misc{li2026provg,
      title={ProVG: Progressive Visual Grounding via Language Decoupling for Remote Sensing Imagery}, 
      author={Ke Li and Ting Wang and Di Wang and Yongshan Zhu and Yiming Zhang and Tao Lei and Quan Wang},
      year={2026},
      eprint={2604.01893},
      archivePrefix={arXiv},
      primaryClass={cs.CV},
      url={https://arxiv.org/abs/2604.01893}, 
}

@misc{jiang2025eaglevisionobjectlevelattributemultimodal,
      title={EagleVision: Object-level Attribute Multimodal LLM for Remote Sensing}, 
      author={Hongxiang Jiang and Jihao Yin and Qixiong Wang and Jiaqi Feng and Guo Chen},
      year={2025},
      eprint={2503.23330},
      archivePrefix={arXiv},
      primaryClass={cs.CV},
      url={https://arxiv.org/abs/2503.23330}, 
}

@inproceedings{yao2026remotereasoner,
  title={Remotereasoner: Towards unifying geospatial reasoning workflow},
  author={Yao, Liang and Liu, Fan and Lu, Hongbo and Zhang, Chuanyi and Min, Rui and Xu, Shengxiang and Di, Shimin and Peng, Pai},
  booktitle={Proceedings of the AAAI Conference on Artificial Intelligence},
  volume={40},
  number={14},
  pages={11883--11891},
  year={2026}
}

@article{liu2025towards,
  title={Towards Faithful Reasoning in Remote Sensing: A Perceptually-Grounded GeoSpatial Chain-of-Thought for Vision-Language Models},
  author={Liu, Jiaqi and Sun, Lang and Fu, Ronghao and Yang, Bo},
  journal={arXiv preprint arXiv:2509.22221},
  year={2025}
}

@misc{wang2025geozero,
      title={GeoZero: Incentivizing Reasoning from Scratch on Geospatial Scenes}, 
      author={Di Wang and Shunyu Liu and Wentao Jiang and Fengxiang Wang and Yi Liu and Xiaolei Qin and Zhiming Luo and Chaoyang Zhou and Haonan Guo and Jing Zhang and Bo Du and Dacheng Tao and Liangpei Zhang},
      year={2025},
      eprint={2511.22645},
      archivePrefix={arXiv},
      primaryClass={cs.CV},
      url={https://arxiv.org/abs/2511.22645}, 
}

@article{zhang2024vision,
  title={Vision-language models for vision tasks: A survey},
  author={Zhang, Jingyi and Huang, Jiaxing and Jin, Sheng and Lu, Shijian},
  journal={IEEE Transactions on Pattern Analysis and Machine Intelligence},
  volume={46},
  number={8},
  pages={5625--5644},
  year={2024},
  publisher={IEEE}
}

@article{zhou2024towards,
  title={Towards vision-language geo-foundation model: A survey},
  author={Zhou, Yue and Zhong, Zhihang and Yang, Xue},
  journal={arXiv preprint arXiv:2406.09385},
  year={2024}
}

@article{wang2025internvl3,
  title={Internvl3. 5: Advancing open-source multimodal models in versatility, reasoning, and efficiency},
  author={Wang, Weiyun and Gao, Zhangwei and Gu, Lixin and Pu, Hengjun and Cui, Long and Wei, Xingguang and Liu, Zhaoyang and Jing, Linglin and Ye, Shenglong and Shao, Jie and others},
  journal={arXiv preprint arXiv:2508.18265},
  year={2025}
}

@inproceedings{soni2025earthdial,
  title={Earthdial: Turning multi-sensory earth observations to interactive dialogues},
  author={Soni, Sagar and Dudhane, Akshay and Debary, Hiyam and Fiaz, Mustansar and Munir, Muhammad Akhtar and Danish, Muhammad Sohail and Fraccaro, Paolo and Watson, Campbell D and Klein, Levente J and Khan, Fahad Shahbaz and others},
  booktitle={Proceedings of the Computer Vision and Pattern Recognition Conference},
  pages={14303--14313},
  year={2025}
}

@article{lan2025text4seg++,
  title={Text4seg++: Advancing image segmentation via generative language modeling},
  author={Lan, Mengcheng and Chen, Chaofeng and Xu, Jiaxing and Li, Zongrui and Ke, Yiping and Jiang, Xudong and Yu, Yingchen and Zhao, Yunqing and Bai, Song},
  journal={arXiv preprint arXiv:2509.06321},
  year={2025}
}

@article{zhan2023rsvg,
  title={Rsvg: Exploring data and models for visual grounding on remote sensing data},
  author={Zhan, Yang and Xiong, Zhitong and Yuan, Yuan},
  journal={IEEE transactions on geoscience and remote sensing},
  volume={61},
  pages={1--13},
  year={2023},
  publisher={IEEE}
}

@article{han2024parameter,
  title={Parameter-efficient fine-tuning for large models: A comprehensive survey},
  author={Han, Zeyu and Gao, Chao and Liu, Jinyang and Zhang, Jeff and Zhang, Sai Qian},
  journal={arXiv preprint arXiv:2403.14608},
  year={2024}
}

@article{wang2025parameter,
  title={Parameter-efficient fine-tuning in large language models: a survey of methodologies},
  author={Wang, Luping and Chen, Sheng and Jiang, Linnan and Pan, Shu and Cai, Runze and Yang, Sen and Yang, Fei},
  journal={Artificial Intelligence Review},
  volume={58},
  number={8},
  pages={227},
  year={2025},
  publisher={Springer}
}

@article{yuan2024rrsis,
  title={Rrsis: Referring remote sensing image segmentation},
  author={Yuan, Zhenghang and Mou, Lichao and Hua, Yuansheng and Zhu, Xiao Xiang},
  journal={IEEE Transactions on Geoscience and Remote Sensing},
  volume={62},
  pages={1--12},
  year={2024},
  publisher={IEEE}
}

@misc{he2026far,
      title={How Far Can Unsupervised RLVR Scale LLM Training?}, 
      author={Bingxiang He and Yuxin Zuo and Zeyuan Liu and Shangziqi Zhao and Zixuan Fu and Junlin Yang and Cheng Qian and Kaiyan Zhang and Yuchen Fan and Ganqu Cui and Xiusi Chen and Youbang Sun and Xingtai Lv and Xuekai Zhu and Li Sheng and Ran Li and Huan-ang Gao and Yuchen Zhang and Bowen Zhou and Zhiyuan Liu and Ning Ding},
      year={2026},
      eprint={2603.08660},
      archivePrefix={arXiv},
      primaryClass={cs.LG},
      url={https://arxiv.org/abs/2603.08660}, 
}

@misc{bai2025qwen3vltechnicalreport,
      title={Qwen3-VL Technical Report}, 
      author={Shuai Bai and Yuxuan Cai and Ruizhe Chen and Keqin Chen and Xionghui Chen and Zesen Cheng and Lianghao Deng and Wei Ding and Chang Gao and Chunjiang Ge and Wenbin Ge and Zhifang Guo and Qidong Huang and Jie Huang and Fei Huang and Binyuan Hui and Shutong Jiang and Zhaohai Li and Mingsheng Li and Mei Li and Kaixin Li and Zicheng Lin and Junyang Lin and Xuejing Liu and Jiawei Liu and Chenglong Liu and Yang Liu and Dayiheng Liu and Shixuan Liu and Dunjie Lu and Ruilin Luo and Chenxu Lv and Rui Men and Lingchen Meng and Xuancheng Ren and Xingzhang Ren and Sibo Song and Yuchong Sun and Jun Tang and Jianhong Tu and Jianqiang Wan and Peng Wang and Pengfei Wang and Qiuyue Wang and Yuxuan Wang and Tianbao Xie and Yiheng Xu and Haiyang Xu and Jin Xu and Zhibo Yang and Mingkun Yang and Jianxin Yang and An Yang and Bowen Yu and Fei Zhang and Hang Zhang and Xi Zhang and Bo Zheng and Humen Zhong and Jingren Zhou and Fan Zhou and Jing Zhou and Yuanzhi Zhu and Ke Zhu},
      year={2025},
      eprint={2511.21631},
      archivePrefix={arXiv},
      primaryClass={cs.CV},
      url={https://arxiv.org/abs/2511.21631}, 
}

@misc{fiaz2025geovlm,
      title={GeoVLM-R1: Reinforcement Fine-Tuning for Improved Remote Sensing Reasoning}, 
      author={Mustansar Fiaz and Hiyam Debary and Paolo Fraccaro and Danda Paudel and Luc Van Gool and Fahad Khan and Salman Khan},
      year={2025},
      eprint={2509.25026},
      archivePrefix={arXiv},
      primaryClass={cs.CV},
      url={https://arxiv.org/abs/2509.25026}, 
}

@misc{min2026remoteshield,
      title={RemoteShield: Enable Robust Multimodal Large Language Models for Earth Observation}, 
      author={Rui Min and Liang Yao and Shiyu Miao and Shengxiang Xu and Yuxuan Liu and Chuanyi Zhang and Shimin Di and Fan Liu},
      year={2026},
      eprint={2604.17243},
      archivePrefix={arXiv},
      primaryClass={cs.CV},
      url={https://arxiv.org/abs/2604.17243}, 
}

@article{zhang2025geor1,
   title={Geo-R1: Improving few-shot geospatial referring expression understanding with reinforcement fine-tuning},
   volume={237},
   ISSN={0924-2716},
   url={http://dx.doi.org/10.1016/j.isprsjprs.2026.04.023},
   DOI={10.1016/j.isprsjprs.2026.04.023},
   journal={ISPRS Journal of Photogrammetry and Remote Sensing},
   publisher={Elsevier BV},
   author={Zhang, Zilun and Guan, Zian and Zhao, Tiancheng and Shen, Haozhan and Cai, Yuxiang and Su, Zhonggen and Shang, Yongheng and Liu, Zhaojun and Yin, Jianwei and Li, Xiang},
   year={2026},
   month=July, pages={113–129} }

@misc{li2026georeason,
      title={GeoReason: Aligning Thinking And Answering In Remote Sensing Vision-Language Models Via Logical Consistency Reinforcement Learning}, 
      author={Wenshuai Li and Xiantai Xiang and Zixiao Wen and Guangyao Zhou and Ben Niu and Feng Wang and Lijia Huang and Qiantong Wang and Yuxin Hu},
      year={2026},
      eprint={2601.04118},
      archivePrefix={arXiv},
      primaryClass={cs.CV},
      url={https://arxiv.org/abs/2601.04118}, 
}

@misc{sun2026geosolver,
      title={GeoSolver: Scaling Test-Time Reasoning in Remote Sensing with Fine-Grained Process Supervision}, 
      author={Lang Sun and Ronghao Fu and Zhuoran Duan and Haoran Liu and Xueyan Liu and Bo Yang},
      year={2026},
      eprint={2603.09551},
      archivePrefix={arXiv},
      primaryClass={cs.CV},
      url={https://arxiv.org/abs/2603.09551}, 
}

@inproceedings{GeoMag,
author = {Ma, Xianzhi and Li, Jianhui and Pei, Changhua and Liu, Hao},
title = {GeoMag: A Vision-Language Model for Pixel-level Fine-Grained Remote Sensing Image Parsing},
year = {2025},
booktitle = {Proceedings of the 33rd ACM International Conference on Multimedia},
pages = {5441–5450},
}

@article{yao2026remoteagent,
  title={Remoteagent: Bridging vague human intents and earth observation with rl-based agentic mllms},
  author={Yao, Liang and Xu, Shengxiang and Liu, Fan and Zhang, Chuanyi and Yao, Bishun and Min, Rui and Li, Yongjun and Ouyang, Chaoqian and Di, Shimin and Zhang, Min-Ling},
  journal={arXiv preprint arXiv:2604.07765},
  year={2026}
}

@misc{ma2026weaveearth,
      title={WeaveEarth: Structured Evidence Construction and Reasoning for Training-Free UHR Remote Sensing Understanding}, 
      author={Xianzhi Ma and Shujun Wang and Xiaohan Li and Hao Liu and Changhua Pei and Jianhui li},
      year={2026},
      eprint={2607.10120},
      archivePrefix={arXiv},
      primaryClass={cs.CV},
      url={https://arxiv.org/abs/2607.10120}, 
}
